\documentclass{article} 
\usepackage{iclr2022_conference} 

\usepackage{amsmath,amsfonts,bm}

\def\eqref#1{equation~\ref{#1}}

\def\1{\bm{1}}

\DeclareMathAlphabet{\mathsfit}{\encodingdefault}{\sfdefault}{m}{sl}
\SetMathAlphabet{\mathsfit}{bold}{\encodingdefault}{\sfdefault}{bx}{n}

\usepackage{hyperref}
\usepackage{url}
\usepackage{booktabs}
\usepackage{graphics}
\usepackage{tabularx}
\usepackage{multirow}
\usepackage[inline]{enumitem}

\title{Towards Clear Expectations for Uncertainty Estimation}

\author{Victor Bouvier, Simona Maggio, Alexandre Abraham \& L\'eo Dreyfus-Schmidt \\ Dataiku\\
\texttt{\{name.surname\}@dataiku.com} 
}

\iclrfinalcopy 
\begin{document}

\maketitle

\begin{abstract}
If Uncertainty Quantification (UQ) is crucial to achieve trustworthy Machine Learning (ML), most UQ
methods suffer from disparate and inconsistent evaluation protocols. We claim this inconsistency results from the unclear requirements the community expects from UQ. This opinion paper offers a new perspective by specifying those requirements through five downstream tasks where we expect uncertainty scores to have substantial predictive power. We design these downstream tasks carefully to reflect real-life usage of ML models. On an example benchmark of 7 classification datasets, we did not observe statistical superiority of state-of-the-art intrinsic UQ methods against simple baselines. We believe that our findings question the very rationale of why we quantify uncertainty and call for a standardized protocol for UQ evaluation based on metrics proven to be relevant for the ML practitioner.
\end{abstract}

\section{Introduction}
\textit{Uncertainty Quantification} (UQ) is a critical requirement for trustworthy Machine Learning (ML) in high-risk applications \citep{hullermeier2021aleatoric}. However the very concept of uncertainty has no agreed upon formal definition. A myriad of definitions and quantifiers for uncertainty exist based on different paradigms. For instance, calibration aims to minimize the \textit{Expected Calibration Error} (ECE) \citep{naeini2015obtaining} or the \textit{Brier Score} \citep{brier1950verification}, Conformal Prediction seeks to produce the smallest set of predictions \citep{angelopoulos2020uncertainty, Angelopoulos2022}, while intrinsic UQ measures the ability to separate uncertainty sources, whether it comes from the data (aleatoric) or the model (epistemic) \citep{Malinin2020, hullermeier2021aleatoric}. 

If those methods cater to different needs, we found the requirements for the UQ they represent not well specified and often heterogeneous, as supported by Table \ref{tab:sota_evaluation}. We initiate to fill this gray area of the literature by proposing clear expectations for UQ through a collection of downstream tasks that  reflect realistic cases the practitioner encounters when needing to deploy an ML model reliably. Crucially, these downstream tasks enable comparison and validation of UQ methods. To this purpose, the contributions of this opinion paper are: 

\begin{itemize}
    \item A set of natural downstream UQ tasks where uncertainty is expected to play a decisive role, such as classification with rejection, error detection, out-of-distribution (OoD) detection, shift detection and performance drop prediction.
    
    \item A methodology to evaluate the predictive power of uncertainty scores on those downstream tasks. To assess the usefulness of uncertainty scores, we always propose simple baselines for the task at hand.    
   \item An example of the above methodology through a benchmark of UQ methods on 7 tabular classification datasets, an under-explored modality for UQ. This shows that, quite surprisingly, for any task simple baselines are often sufficient alternatives to more advanced uncertainty quantification.
 \end{itemize}

\section{Challenges in Uncertainty Evaluation}
\label{sec:challenges}

 While prior works in UQ propose a plethora of different techniques and specific tasks to evaluate them, there is no shared requirement an uncertainty score should fulfill. We outline the diverse landscape of UQ evaluation and highlight the requirements we believe an uncertainty estimation should comply with to be used to guarantee trustworthy ML.

\paragraph{UQ Evaluation in the literature.}

Many methods has been designed for uncertainty estimation \citep{Eyke2020, Abdar2021}, yet only limited work has focused on how to evaluate and benchmark these approaches \citep{Nado2021, Malinin2021}, mainly focusing on deep classification of images \citep{Filos2019, Atanov2020}.
As summarized in Table \ref{tab:sota_evaluation}, research papers proposing new UQ estimation methods do not share the same evaluation protocol, some focusing on calibration metrics, some on downstream tasks metrics. Error and Out-of-Distribution (OoD) detection are the most common downstream tasks, but the validation remains often partial with either a single task evaluated or a qualitative assessment that uncertainty estimates increase in specific situations \citep{Ovadia2019}. Importantly, the lack of comparison to a  baseline tailored to the task at hand makes it difficult to assess the actual practicality of an uncertainty score compared to a simpler alternative, such as using an anomaly detection method for the OoD detection task. Additionally, designing downstream tasks is problematic too. From defining what, an OoD sample is, for which there is no standard in particular for tabular data, to comparing performances on the error detection task that could be misleading \citep{Atanov2020}.

\begin{table}[t]
\scriptsize
\def \colone{2.7cm}
\begin{tabularx}{\textwidth}{m{2.3cm}m{3.2cm}m{3.2cm}X}
\toprule
Reference  & Dataset & OOD Dataset  & Downstream tasks \\
\toprule
\multirow{2}{\colone}{\cite{Lakshminarayanan2017}}
    & UCI Regression & No OOD & Calibration \\   & MNIST, SVHN, ImageNet & Left-out class & Calibration, OoD detection \\
\midrule
\multirow{3}{\colone}{\cite{Ovadia2019}} & MNIST & Rotated, Translated, NotMNIST & \multirow{3}{*}{Calibration} \\
    & CIFAR10, ImageNet & Gaussian blurred, SVHN &  \\
    & 20 NG (even classes) & 20 NG odd classes, One billion word &  \\
\midrule 
\multirow{2}{\colone}{\cite{Filos2019}}
    & UCI Regression & No OOD & Calibration \\ 
    & Diabete retinopathy & APTOS 2019 blindness detection & Retention \\ 
\midrule
\multirow{2}{\colone}{\cite{Malinin2020}}          & UCI regression-classification & Year MSD                         & Retention, OoD detection\\
\midrule
\multirow{2}{\colone}{\cite{shaker2020aleatoric}}
    & spect &  \multirow{2}{*}{No OOD}  & \multirow{2}{*}{Retention} \\
    & diabetes  &  &                     \\
\midrule
\cite{Nado2021} & ImageNet & ImageNet-C/A/V2 & Calibration, OoD detection\\
\midrule
\cite{daxberger2021laplace} & WILDS & WILDS shifts & Calibration \\
\midrule
\multirow{2}{\colone}{\cite{Malinin2021}}
    & weather ante 2019-04, not snow & weather post 2019-07, snow & Retention \\   
    & Machine translation EN-RU newspaper & Other vocabulary (Reddit) &   \\
\bottomrule
\end{tabularx}
\caption{Evaluation Metrics and Tasks in SOTA works on UQ. Prior works in the literature do not share a standard evaluation protocol, using different datasets or metrics, making it difficult to assess the actual progress of Uncertainty Quantification for Machine Learning.}
\label{tab:sota_evaluation}
\end{table}

\paragraph{Requirements for reliable ML.}

We believe that robust uncertainty estimation is a cornerstone towards trustworthy ML that naturally leads to the following five downstream tasks where a good uncertainty score should perform well.

\textit{Retention.} Uncertainty scores should help deferring the prediction to a human expert, which is a typical case of medical diagnosis with ML where difficult samples are deferred to the medical expert. Formally, for a budget $r \in [0,1]$, an oracle provides the ground-truth for $r\%$ of the data with higher uncertainty score. As a metric, we report the area under the $F1$-score between oracle-augmented predictions and the true label, when varying $r\in [0,1]$. 

\textit{Error detection.} Uncertainty scores should help detecting model errors, as higher uncertainty should correlate with higher error probability. It reflects the situation where model errors are highly detrimental, and one should rather abstain from predicting. As a metric, we report the Area Under the ROC curve (AUROC) when the uncertainty score is used as the prediction score of the test set errors. 

\textit{Out-of-Distribution (OoD) detection.} We expect uncertainty scores to support detection of OoD samples from different domains, that are generated via splits along specific feature values. As the model is expected to be more uncertain on OoD samples, we report the AUROC when the uncertainty score is used as the prediction score of the test set errors to detect the domain, \textit{i.e.} whether the sample comes from the shifted or unshifted dataset. 

\textit{Shift detection.} Uncertainty scores are used as proxy for detecting data shift \citep{rabanser2019failing}. In practice, we perform a Kolmogorov-Smirnov test between the distribution of uncertainty scores on the test set and of shifted datasets, also generated via splits along specific feature values. We report the accuracy metric of detecting shift with a signifiance level $\alpha=0.05$ based on 100 bootstrap of shifted datasets.

\textit{Performance Drop Prediction.} We expect that the ratio of uncertainty scores above a given threshold reflect the error rate on several shifted datasets, generated as in the task above. The threshold is selected so that the ratio of samples with uncertainty above the threshold matches the error rate on the test set, 
similarly to the Averaged Thresholded Confidence (ATC) technique \citep{Garg2022}. We report as metric the mean absolute error between the estimated performance drop and the true performance drop. 

It is worth noting the described tasks share similarities. Both retention and error detection assess how valuable is an uncertainty score for detecting difficult in-distributions samples, \textit{i.e.} for which the model is likely to fail to predict, and depend on the primary model.  Both OoD detection and shift detection assess how valuable is an uncertainty score for detecting samples that are far from the data distribution the model was trained on. The task of the performance drop prediction combines both assessments and is inspired by recent work \citep{Garg2022}. 

\section{Benchmarking UQ Methods for Reliable ML}

 \begin{table}[t]
 \resizebox{\columnwidth}{!}{
 \begin{tabular}{lllccccc}
\toprule
                         Primary model &   Agnostic UQ &           Score &       Retention & Error Detection & OoD Detection & Shift Detection & Perf. Drop Pred. \\




\toprule

Logistic Regression & Isotonic Calibration &  Max-Confidence &  \textbf{0.942} &  \textbf{0.763} &             0.460 &           0.489 &       \textbf{0.096} \\
\cmidrule{2-8}
 &  CP (Confidence)  &          p-value &           0.937 &           0.719 &             0.486 &           0.537 &       \textbf{0.075} \\
&  &     Credibility &           0.929 &           0.665 &             0.480 &           0.409 &       \textbf{0.088} \\
&  &      Confidence &           0.929 &           0.658 &             0.466 &           0.355 &       \textbf{0.091} \\
\cmidrule{2-8}
&  None &  Baseline &  \textbf{0.942}${}^\star$ &  \textbf{0.768}${}^\star$ &             \textbf{0.639}${}^\dag$ &       \textbf{0.956}${}^\dag$ &       \textbf{0.080}${}^\star$ \\

\toprule 

Random Forest & Isotonic Calibration &  Max-Confidence & \textbf{0.953} &  \textbf{0.789} &             0.560 &           0.877 &                0.053 \\
\cmidrule{2-8}
 &  CP (Confidence)  &          p-value &          0.946 &           0.738 &             0.576 &           0.847 &       \textbf{0.036} \\
&  &     Credibility &         0.941 &           0.694 &             0.572 &           0.812 &       \textbf{0.041}  \\
&  &      Confidence &          0.941 &           0.697 &             0.567 &           0.836 &       \textbf{0.046} \\
\cmidrule{2-8}
&  None &  Baseline &  \textbf{0.953}${}^\star$ &  \textbf{0.797}${}^\star$ &             \textbf{0.639}${}^\dag$ &       \textbf{0.956}${}^\dag$ &       \textbf{0.035}${}^\star$ \\

\toprule 

Gradient Boosting Trees & Isotonic Calibration &  Max-Confidence & \textbf{0.951} &  \textbf{0.785} &             0.506 &           0.811 &                0.032  \\
\cmidrule{2-8}
 &  CP (Confidence)  &          p-value &           0.945 &           0.736 &             0.512 &           0.796 &       \textbf{0.018} \\
&  &     Credibility &         0.940 &           0.695 &             0.510 &           0.782 &                0.028  \\
&  &      Confidence &          0.940 &           0.695 &             0.503 &           0.759 &                0.028 \\
\cmidrule{2-8}
&  None &  Baseline & \textbf{0.951}${}^\star$ &  \textbf{0.794}${}^\star$  &             \textbf{0.639}${}^\dag$ &       \textbf{0.956}${}^\dag$ &       \textbf{0.016}${}^\star$ \\

\toprule 

Multi-Layers Perceptron & Isotonic Calibration &  Max-Confidence & \textbf{0.953} &  \textbf{0.789} &             0.459 &           0.812 &       \textbf{0.091}  \\
\cmidrule{2-8}
 &  CP (Confidence)  &          p-value &       0.945 &           0.718 &             0.461 &           0.761 &       \textbf{0.080}  \\
&  &     Credibility &         0.942 &           0.694 &             0.459 &           0.748 &       \textbf{0.080} \\
&  &      Confidence &        0.942 &           0.692 &             0.453 &           0.717 &       \textbf{0.082} \\
\cmidrule{2-8}
&  None &  Baseline &\textbf{0.953}${}^\star$ &  \textbf{0.793}${}^\star$ &             \textbf{0.639}${}^\dag$ &       \textbf{0.956}${}^\dag$ &       \textbf{0.079}${}^\star$  \\

\toprule

Deep Ensemble & None & Total &  \textbf{0.955} &  \textbf{0.798} &             0.506 &           0.899 &       \textbf{0.056} \\
 (Intrinsic UQ) &  &       Aleatoric &  \textbf{0.955} &  \textbf{0.796} &             0.470 &           0.893 &                0.092 \\
&  &       Epistemic &           0.947 &           0.748 &    0.579 &  0.932 &                0.092 \\
\cmidrule{2-8}
& Isotonic Calibration &  Max-Confidence &  \textbf{0.955} &          \textbf{0.792} &             0.500 &           0.824 &       \textbf{0.070} \\
\cmidrule{2-8}
& CP (Confidence) &          p-value &           0.947 &           0.725 &             0.506 &           0.749 &       \textbf{0.056} \\
&  &     Credibility &           0.944 &           0.698 &             0.503 &          0.734 &       \textbf{0.059} \\
&  &      Confidence &           0.944 &           0.695 &             0.495 &            0.721  &       \textbf{0.060} \\
\cmidrule{2-8}
&  None &  Baseline & \textbf{0.955}${}^\star$ &  \textbf{0.798}${}^\star$ &             \textbf{0.639}${}^\dag$ &       \textbf{0.956}${}^\dag$ &       \textbf{0.056}${}^\star$  \\

\toprule
CatBoost & None &           Total &  \textbf{0.957} &  \textbf{0.806} &             0.512 &  0.894 &       \textbf{0.046} \\
(Intrinsic UQ) &  &       Aleatoric &  \textbf{0.957} &  \textbf{0.806} &             0.512 & 0.874 &       \textbf{0.046} \\
&  &  Epistemic &           0.939 &           0.676 &    0.551 &  0.895 &                0.080 \\
\cmidrule{2-8}
& Isotonic Calibration & Max-Confidence &  \textbf{0.956} & 0.796 &             0.505 &           0.818 &       \textbf{0.066} \\
\cmidrule{2-8}
& CP (Confidence) &          p-value &         0.950 &           0.740 &         0.537 & 0.773  &       \textbf{0.046} \\
&  &     Credibility &          0.946 &           0.706 &             0.510 &          0.769 &       \textbf{0.052} \\
&  &      Confidence &          0.946 &           0.709 &             0.502 &          0.782 &       \textbf{0.054} \\

\cmidrule{2-8}
&  None &  Baseline &  \textbf{0.957}${}^\star$ &  \textbf{0.806}${}^\star$ &             \textbf{0.639}${}^\dag$ &       \textbf{0.956}${}^\dag$ &        \textbf{0.046}${}^\star$  \\
\bottomrule
\end{tabular}

 }
 \caption{Results of the example 7 datasets benchmark. In bold, metrics with overlapping confidence interval with best performer based on statistics over 10 different seeds, as described in \textbf{Experimental setup}. We observe strong performance from the simple baselines, whether for error detection or OoD and shift detection ($\dag$), questioning the relative benefit of specific UQ methods.}
 \label{table:summary_experiments}
 \end{table}

As a validation protocol, we use uncertainty score derived from UQ methods on the five downstream tasks together with a simple baseline to assess the actual practicality of an uncertainty score.

\paragraph{Comparing UQ.} We aim to compare both agnostic UQ, \textit{i.e.} quantifying uncertainty of a black-box model, and intrinsic UQ, \textit{i.e.} a model that natively quantifies its uncertainty. First, we study two agnostic UQ; \textit{Isotonic Calibration} (IC) and \textit{Conformal Prediction}\footnote{We used the confidence as non-conformity score. We refer the reader to  \citep{angelopoulos2020uncertainty} for a comprehensive overview of conformal predictions.} (CP). Various uncertainty scores can be derived from conformal predictions; the $\textit{confidence}$ which measures how certain the model is that the prediction set is a singleton, the \textit{credibility} which measures how certain the model is that the prediction set is not empty,  and the $\textit{p-value}$ that is the quantile of the non-conformity score associated with prediction in the calibration samples with label equal to the prediction \citep{Shafer2008}.  For the isotonic calibration module, we use as uncertainty score the \textit{Max-Confidence} uncertainty score as 1 minus the maximal predicted probability over classes. Second, we study intrinsic UQ that aims to separate the aleatoric from the epistemic uncertainty. To this purpose, we compare three different scores: the total, aleatoric and epistemic uncertainties \citep{hullermeier2021aleatoric}. To compare as fairly as possible intrinsic and agnostic UQ, we fix the base model of intrinsic UQ, considering it as a black-box model  on which we plug agnostic UQ. Such methodology is motivated by the conclusion from \citep{Atanov2020} which stresses that error-based tasks (in our particular setup: retention, error detection and performance drop prediction) depend on the primary model, thus are not directly comparable across different primary models.

\paragraph{Baseline.} To assess the utility of an uncertainty score in a downstream task, we compare it against simpler alternatives. For the error-based tasks that are model dependent, we used the \textit{Max-Confidence} score from the black-box primary model, \textit{i.e.} without any UQ. In table \ref{table:summary_experiments}, we flag results using this score with $\star$ symbol. For OoD-based tasks, we report a naive anomaly detection score based on the mean distance of a sample to its ten nearest neighbors with same label on training data building an uncertainty score which is independent on the primary model. Being model independent, it  is worth noting the results based on the anomaly detection score are shared for all models.  In table \ref{table:summary_experiments}, we flag results using this score with $\dag$ symbol.

\paragraph{Experimental setup.}
We focus on binary classification tasks from seven tabular datasets. We used the UCI datasets\footnote{\url{archive.ics.uci.edu/ml/datasets}} \textit{Adult Income}, \textit{Video Games}, \textit{Default Credit Card}, \textit{Bank}, \textit{Heart}, \textit{Forest Covertype} and \textit{BNG Zoo}\footnote{For datasets that are not binary tasks (Forest Covertype and BNG Zoo), we derive a binary task in a one-vs-all fashion using the majority class as the positive class.} We detail in Appendix \ref{section:ood_data} how OoD data is obtained.  For agnostic UQ, we studied four different primary models: Logistic Regression, Random Forest, Gradient Boosting Trees and a Multi-Layer Perceptron (MLP) implemented using the \texttt{scikit-learn} library \citep{pedregosa2011scikit} with default parameters for each estimator. For intrinsic UQ, we used two influential methods: \textit{Deep Ensemble} \citep{Lakshminarayanan2017} with an ensemble of 10 MLPs, and \textit{CatBoost} \citep{Malinin2020} with $\mathtt{RMSEWithUncertainty}$ loss.  We compute mean metrics $m$ and their standard deviation $\sigma$ over 10 random seeds and aggregated on the seven datasets. Bold metrics are such that their confidence interval $[m - \sigma, m+\sigma]$ lower bound is higher than any other confidence interval higher bound. As error-based tasks (retention, error detection and performance drop prediction) are not directly comparable \citep{Atanov2020}, we bold metrics for each primary model separately.

\paragraph{Analysis.}

 Although primary models are not directly comparable on error-based tasks, results presented in Table \ref{table:summary_experiments} show that metrics range in a similar way. More importantly, we find that no UQ method clearly outperforms others and that tailored baselines performs similarly on error-based tasks than UQ methods, and even outperforms them on OoD detection and shift detection. Thus, if the utility of UQ methods for trustworthy ML is to be judged on those downstream tasks, we advocate for the use of those simple baselines.

\section{Conclusion}
We challenge the evaluation of \textit{Uncertainty Quantification} (UQ) as a literature review showed inconsistent evaluation protocols.
For UQ to help guarantee more reliable use of ML models, we proposed five natural downstream tasks. To our knowledge, this new perspective for UQ evaluation is the first that allows to compare very different paradigms for UQ, from Conformal Prediction to Intrinsic UQ. Quite surprisingly, our example benchmark shows that relying simply on the probability estimates of the primary model is a fair, simple and robust baseline for error-related downstream tasks, while for OoD-related tasks more complex techniques based on separation of sources of uncertainties perform significantly lower than a simple baseline. As concluding words, we draw recommendation for future work towards more reliable UQ evaluation protocol;
\begin{enumerate}
    \item Adding more datasets would strengthen the statistical significance of results,  and allow the use of more advanced ranking methods, such as autorank \citep{herbold2020autorank}.
    \item Developing more downstream tasks that captures the expectation from uncertainty scores together with simple generic baseline when available.
    \item Our analysis is restricted to binary classification task and would benefit being extended to regression tasks.
\end{enumerate}

\bibliography{iclr2022_conference}
\bibliographystyle{iclr2022_conference}

\appendix
\section{Out-of-Distribution data}
\label{section:ood_data}
We detail how we split the dataset to obtain \textit{Out-of-Distribution} (OoD) data for our experiments; 
\begin{itemize}
    \item Adults income: we split in-distribution data and OoD data with respect to the "gender" features; "male" is in-distribution, "female" is out-of distribution.
    \item Video games: we split in-distribution data and OoD data with respect to the "Genre" features; "Action" is OoD, other values are in-distribution.
    \item Heart: we split in-distribution data and OoD data with respect to the "gender" features; "2" is out-of-distribution, other values are in-distribution.
    \item Bank: we split in-distribution data and OoD data with respect to the "job" features; "student" is OoD, other values are in-distribution.
    \item Default of credit card client: we split in-distribution data and OoD data with respect to the "SEX" features; "1" is OoD, other values are in-distribution.
    \item Forest Covertype: we split in-distribution data and OoD data with respect to the "Wilderness\_Area1" features; "1" is out-of-distribution, other values are in-distribution.
    \item BNG Zoo: we split in-distribution data and OoD data with respect to the "domestic" features; "True" is OoD, other values are in-distribution.
\end{itemize}
We then remove the feature used for the split for both the in-distribution and the out-of-distribution datasets.
Shifted data for the shift detection and performance drop prediction tasks should be generated via synthetic perturbations as outlined in Section \ref{sec:challenges}. We plan to include synthetically shifted data in the benchmark as future work, but in the experiments presented in the paper we used the OoD dataset as a shifted dataset for shift detection and performance drop prediction as well. 

\end{document}